\documentclass[10pt,twocolumn,letterpaper]{article}

\usepackage{iccv}
\usepackage{times}
\usepackage{epsfig}
\usepackage{graphicx}
\usepackage{amsmath}
\usepackage{amssymb}
\usepackage[american]{babel}
\usepackage{microtype}
\usepackage{subfigure} 
\usepackage{slashbox}

\usepackage{enumitem}
\setlist[itemize]{noitemsep, nolistsep}

\usepackage[pagebackref=true,breaklinks=true,letterpaper=true,colorlinks,bookmarks=false]{hyperref}

\iccvfinalcopy 

\def\iccvPaperID{32} 

\ificcvfinal\fi

\begin{document}

\title{Unsupervised Deep Feature Transfer for Low Resolution Image Classification}


\author{Yuanwei Wu\textsuperscript{1}\thanks{This work was done when the first author took internship at MERL.}, Ziming Zhang\textsuperscript{2}\thanks{Corresponding author.}, and Guanghui Wang\textsuperscript{1}\\
\textsuperscript{1} EECS, The University of Kansas, Lawrence, KS 66045\\
\textsuperscript{2} Mitsubishi Electric Research Laboratories (MERL), Cambridge, MA 02139\\
{\tt\small y262w558@ku.edu, zzhang@merl.com, ghwang@ku.edu}
}

\maketitle
\ificcvfinal\thispagestyle{empty}\fi

\begin{abstract}
In this paper, we propose a simple while effective unsupervised deep feature transfer algorithm for low resolution image classification. No fine-tuning on convenet filters is required in our method. We use pre-trained convenet to extract features for both high- and low-resolution images, and then feed them into a two-layer feature transfer network for knowledge transfer. A SVM classifier is learned directly using these transferred low resolution features. Our network can be embedded into the state-of-the-art deep neural networks as a plug-in feature enhancement module. It preserves data structures in feature space for high resolution images, and transfers the distinguishing features from a well-structured source domain (high resolution features space) to a not well-organized target domain (low resolution features space). Extensive experiments on VOC2007 test set show that the proposed method achieves significant improvements over the baseline of using feature extraction. 

\end{abstract}

\section{Introduction}\label{sec:intro}


Recently, deep neural networks have demonstrated impressive results in image classification~\cite{krizhevsky2012imagenet,he2016deep,8794588,xu2019adaptively}, object detection~\cite{girshick2014rich,ren2015faster,ma2018mdcn,zhu2018visdrone}, instance segmentation~\cite{he2017mask}, visual tracking~\cite{zhu2018visdrone,wu2017vision,bharati2018real,bharati2016fast} depth estimation~\cite{he2018learning,he2018spindle}, face recognition~\cite{cen2019dictionary}, and image translation~\cite{isola2017image,xu2019toward,xu2019adversarially,xu2019stacked}. The success of DNNs has become possible mostly due to a large amount of annotated datasets~\cite{deng2009imagenet}, as well as advances in computing resources and better learning algorithms~\cite{goyal2017accurate,Zhang_2018_CVPR}. Most of these works typically assume that the images are of sufficiently high resolution (\eg $224\times224$ or larger).

The limitation of requiring large amount of data to train DNNs has been alleviated by the introduction of transfer learning techniques. A common way to make use of transfer learning in the context of DNNs is to start from a pre-trained model in a similar task or domain, and then finetune the parameters to the new task. For example, the pre-trained model on ImageNet for classification can be finetuned for object detection on Pascal VOC~\cite{girshick2014rich,ren2015faster}.

\begin{figure}[t]
	\begin{center}
	\includegraphics[width=\linewidth]{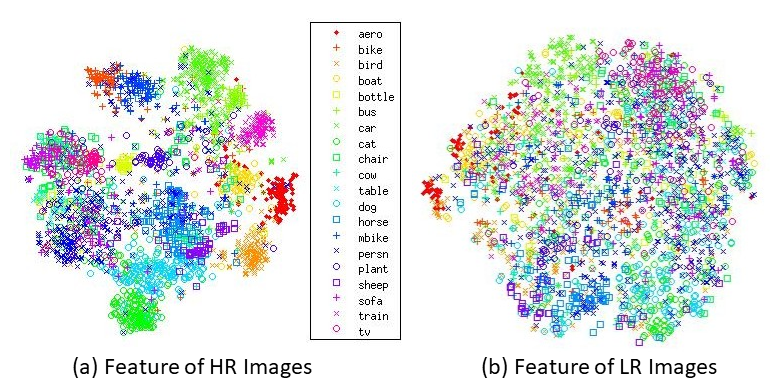}
	\end{center}
	\vspace{-8pt}
	\caption{The tSNE~\cite{maaten2008visualizing} of deep features (2048-D) of VOC2007 train set extracted from pool5 layer of pre-trained resnet-101~\cite{he2016deep}. (a) Feature of High Resolution (HR) images, and (b) feature of Low Resolution images. The HR features are well separated, however, the LR features are mixed together.}
	\label{fig:motivation_figure}
\end{figure}

In this paper, we focus on low resolution (\eg $32\times32$ or less) image classification as for privacy purpose, it is common to use low resolution images in real-world applications, such as face recognition in surveillance videos~\cite{zou2011very}. Without additional information, learning from low resolution images always reduces to an ill-posed optimization problem, and achieves a much degraded performance~\cite{pinheiro2015learning}. 

As shown in Fig.~\ref{fig:motivation_figure}, the deep feature of high resolution images extracted from pre-trained convenet has already learned discriminative per-class feature representation. Therefore, it is able to be well separated in the tSNE visualization. However, the extracted feature of low resolution images is mixed together. A possible solution is to exploit the transfer learning, leveraging the discriminative feature representation from high resolution images to low resolution images.

\begin{figure*}[t]
	\begin{center}		
		\includegraphics[width=\linewidth]{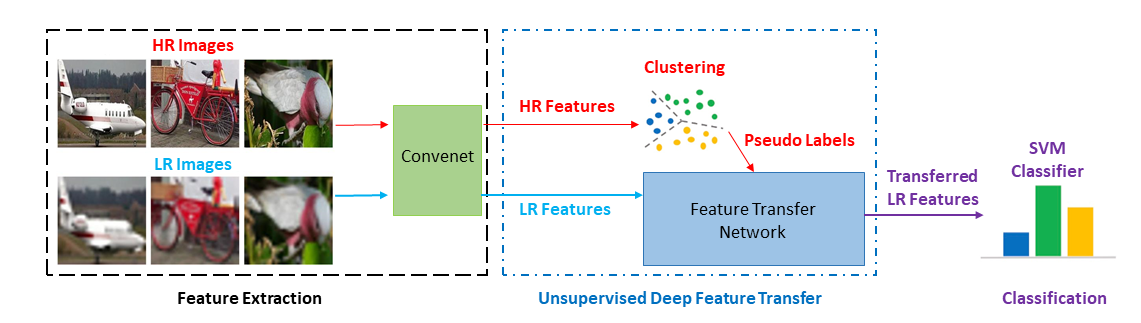}
	\end{center}
	\vspace{-8pt}
	\caption{The overview of proposed unsupervised deep feature transfer algorithm. It consists of three modules. In the feature extraction module, a pre-trained deep convenet is used as feature extractor to obtain HR and LR features from HR and LR images, respectively. Then, we cluster the HR features to obtain pseudo-labels, which are used to guide the feature transfer learning of LR features in the feature transfer network. Finally, a SVM classifier is trained on the transferred LR features.}
	\label{fig:Illustration_of_our_method}
\end{figure*}

In this paper, we propose a simple while effective unsupervised deep feature transfer approach that boosts classification performance in low resolution images. We assume that we have access to high resolution labeled images during training, but at test we only have low resolution images. Most existing datasets are high resolution. Moreover, it is much easier to label subcategories in high resolution images. Therefore, we believe it is a reasonable assumption. We aim to transfer knowledge from such high resolution images to real world scenarios that only have low resolution images. The basic intuition behind our approach is to utilize high quality discriminative representations in the training domain to guide feature learning for the target low resolution domain.



The contributions of our work have three-fold. 
\begin{itemize}
\item No fine-tuning on convenet filters is required in our method. We use pre-trained convenet to extract features for both high resolution and low resolution images, and then feed them into a two-layer feature transfer network for knowledge transfer. A SVM classifier is learned directly using these transferred low resolution features. Our network can be embedded into the state-of-the-art DNNs as a plug-in feature enhancement module. 
\item It preserves data structures in feature space for high resolution images, by transferring the discriminative features from a well-structured source domain (high resolution features space) to a not well-organized target domain (low resolution features space).
\item Our performance is better than that of baseline using feature extraction approach for low resolution image classification task. 
\end{itemize}

\section{Related Work}
\label{sec:RelatedWork}
Our method is closely related to unsupervised learning of features and transfer learning.

\textbf{Unsupervised learning of features:} Clustering has been widely used for image classification~\cite{caron2018deep,yang2016joint,ji2018invariant}. Ji~\etal~\cite{ji2018invariant} propose invariant information clustering relying on statistical learning by optimising mutual information between related pairs for unsupervised image classification and segmentation. Caron~\etal~\cite{caron2018deep} present a clustering method that jointly learns the parameters of a neural network and the cluster assignments of the resulting features. Yang~\etal~\cite{yang2016joint} propose an approach to jointly learn deep representations and image clusters by combining agglomerative clustering with CNNs and formulate them as a recurrent process.

\textbf{Transfer learning:} It is commonly used in the scenario where the training and testing data distributions are different. 
Saenko~\etal~\cite{saenko2010adapting} learn a regularized non-linear transformation in the context of object recognition to minimize the effect of domain-induced changes in the feature distribution. Chen~\etal~\cite{chen2015net2net} transfer knowledge stored in one previous network into each new deeper or wider network to accelerate the training of a significantly larger neural network. Yosinski~\etal~\cite{yosinski2014transferable} experimentally study the transferability of hierarchical features in deep neural networks. Azizpour~\etal~\cite{azizpour2016factors} investigate the factors of transferability of a generic deep convolutional networks such as the network architecture, distribution of the training data, etc. Tzeng~\etal~\cite{tzeng2015simultaneous} learn a CNN architecture to optimize domain invariance and transfer information between tasks. Long~\etal~\cite{long2015learning} propose a deep adaptation network architecture to match the mean embeddings of different domain distributions in a reproducing kernel Hilbert space. Guo~\etal~\cite{guo2019spottune} propose an adaptive fine-tuning approach to find the optimal fine-tuning strategy per instance for the target data. Readers can refer to~\cite{pan2010survey} and the references therein for details about transfer learning.

\section{Proposed Approach}\label{sec:Proposed_Approach}
This section describes the proposed unsupervised deep feature transfer approach. 
\subsection{Preliminary}\label{subsec:preliminary}
With the recent success of deep learning in computer vision, the deep convnets have become a popular choice for representation learning, to map raw images to an embedding vector space of fixed dimensionality. In the context of supervised learning, they could achieve better performance than humanbeings on standard classification benchmarks~\cite{he2015delving,krizhevsky2012imagenet} when trained with large amount of labelled data. 

Let $f_\theta$ denote the convenet mapping function, where $\theta$ is the corresponding learnable parameters. We refer to the vector obtained by applying this mapping to an image as feature or features. Given a training set $X=\{x_1, \cdots, x_N\}$ of $N$ images, and the corresponding ground truth labels $Y=\{y_1, \cdots, y_N\}$, we want to find an optimal parameter $\theta^*$ such that the mapping $f_\theta^*$ predicts good general features. Each image $x_i$ associates with a class label $y_i$ in $\{0, 1\}^k$. Let $g_w$ denote a classifier with parameter $\omega$. The classifier would predict the labels on top of the features $f_\theta(x_i)$. The parameter $\theta$ of the mapping function and the parameter $\omega$ of the classifier are then learned jointly by optimizing the following objective function:

\begin{equation}
    \min_{\theta, \omega} \frac{1}{N}\sum_{i=1}^{N} \mathcal{L}(g_w(f_\theta(x_i), y_i))\,,
    \label{equ:multinominal_loss}
\end{equation}
where $\mathcal{L}$ is the multinominal logistic loss for measuring the difference between the predicted labels and ground-truth labels given training data samples.

\subsection{Unsupervised Deep Feature Transfer}\label{sec:Unsupervised_DFT}
The idea of this work is to boost the feature learning for low resolution images by exploiting the capability of unsupervised deep feature transfer from the discriminative high resolution feature. The overview of proposed approach is shown in Fig.~\ref{fig:Illustration_of_our_method}. It consists of three modules: feature extraction, unsupervised deep feature transfer, and classification, discussed below. 

\textbf{Feature extraction.} We observe that the deep features extracted from convenet could generate well separated clusters as shown in Fig.~\ref{fig:motivation_figure}. Therefore, we introduce the transfer learning to boost the low resolution features learning via the supervision from high resolution features. Then, we extract the features (N-Dimensional) of both high and low resolution images from a pre-trained deep convenet. More details are described in Sec.~\ref{sec:Implementation_Details}.

\textbf{Unsupervised deep feature transfer.} We propose a feature transfer network to boost the low resolution features learning. However, in our assumption, the ground truth labels for low resolution images are absent. Therefore, we need to make use of the information from high resolution features. In order to do this, we propose to cluster the high resolution features and use the subsequent cluster assignments as ``pseudo-label" to guide the learning of feature transfer network with low resolution features as input. Without loss of generality, we use a standard clustering algorithm, k-means. The k-means takes a high resolution feature as input, in our case the feature $f_\theta(x_i)$ extracted from the convenet, and clusters them into $k$ distinct groups based on a geometric criterion. Then, the pseudo-label of low resolution features are assigned by finding its nearest neighbor to the $k$ centroids of high resolution features. Finally, the parameter of the feature transfer network is updated by optimizing Eq.~(\ref{equ:multinominal_loss}) with mini-batch stochastic gradient descent. 



\textbf{Classification.} The final step is to train a commonly used classifier such as Support Vector Machine (SVM) using the transferred low resolution features. In testing, given only the low resolution images, first, our algorithm extracts the features. Then feeds them to the learned feature transfer network to obtain the transferred low resolution features. Finally, we run SVM to get the classification results directly.

\begin{table*}[t]
	\centering\footnotesize 
	\setlength\tabcolsep{2.5pt}
	\begin{center}
		\begin{tabular}{|c|c|c|c|c|c|c|c|c|c|c|c|c|c|c|c|c|c|c|c|c|c|}
			\hline
			& aero & bike & bird & boat & bottle & bus & car & cat & chair & cow  & table & dog & horse & mbike & persn & plant & sheep & sofa & train & tv & mAP \\  \hline
			
			Baseline-HR & 97.6 & 92.7 & 89.2 & 85.8 & 90.6 & 87.5& 96.2 & 94.3 & 81.4 & 83.3 & 80.0 & 86.9 & 84.2 & 90.0 & 95.4 & 95.0 & 88.3 & 71.6 & 96.0 & 95.9  & 89.1\\ \hline
			
			Baseline-LR & 87.5 & 84.8 & 77.5 & 77.4 & 80.4 & 76.5& 90.6 & 72.1 & 75.1 & 72.9 & 69.5 & 65.0 & 71.7 & 73.9 & 92.8 & 90.8& 78.3 & 48.6 & 83.3 & 92.3 & 78.1\\ \hline
			
			Ours & 89.1 & 86.5 & 80.1 & 78.1 & 79.6 & 77.4 & 92.4 & 75.4 & 79.4 & 73.2 & 72.5 & 68.5 & 74.0 & 77.1 & 95.0 & 91.9 & 77.6 & 53.4 & 86.1 & 92.5 & 80.0 \\ \hline
		\end{tabular}
	\end{center}
	\caption{Per-class average precision (\%) for object classification on the VOC2007 test set.}
	\label{table:VOC07_Bbox_baseline}
\end{table*}

\section{Experiments}\label{sec:Experiments}
\subsection{Dataset}\label{subsec:dataset}

We conduct the low resolution classification on the PASCAL VOC2007 dataset~\cite{Everingham15} with 20 object classes. There are $5,000$ images in VOC2007 trainval set and $4,952$ images in VOC2007 test set. However, the images in the dataset are high resolution images only. We follow~\cite{lin2014microsoft} to generate the low resolution images. In this work, we generate high resolution images by resizing the original images to $224\times 224$ using bicubic interpolation. We generate the low resolution images by down-sampling the original to $32\times 32$, and then up-sampling to $224\times 224$.

\subsection{Implementation Details}\label{sec:Implementation_Details}
We conduct our experiment using Caffe~\cite{jia2014caffe}. We use the resnet-101~\cite{he2016deep} pre-trained on ILSVRC$2012$\footnote{We download the Caffe Model from \url{https://github.com/BVLC/caffe/wiki/Model-Zoo}}~\cite{russakovsky2015imagenet} as the backbone convenet to extract the features from high and low resolution images. We extract the features from the pool5 layer, which gives a feature vector with dimension of $N=2048$. 

The feature transfer network is a two-layer fully connected network. We conduct grid search to find the optimal design for the network architecture, see Sec.~\ref{sec:feature_feature_network}. It is initialized using MSRA~\cite{jia2014caffe} initialization. We train the feature transfer network using stochastic gradient descent with weight decay $0.0005$, momentum $0.9$, batch size $1,000$, epoch $1,000$, total iteration $31,561$. The initial learning rate is $0.01$, and is decreased by $10$ after every $15,000$ iterations. 

\begin{table}[th]
	\begin{center}\footnotesize
		\begin{tabular}{|c|*{5}{c|}}\hline
			\backslashbox{$N_2$}{$N_1$}
			& 256 & 512 & 1024 & 2048 & 4096\\\hline
			20 & 0.704  & 0.741 & \textbf{0.771}& 0.786 & \textbf{0.800} \\\hline
			100 & 0.718 & \textbf{0.752}  & 0.768 & \textbf{0.789} & \textbf{0.800} \\\hline
			200 & \textbf{0.727} & 0.746 & \textbf{0.771} & 0.788 & \textbf{0.800} \\\hline
			500 & 0.717 & 0.743 & 0.766 &  0.784 & 0.795 \\\hline
			1000 &0.713 & 0.743 & 0.762 & 0.783 & 0.793\\\hline
			2048 & 0.718 & 0.739 & 0.765 & 0.783 & 0.794 \\\hline
		\end{tabular}
	\end{center}
	\caption{We use grid search to find the optimal combination of $N_1$ and $N_2$ for the two-layer feature transfer network by calculating the mean average precision (mAP) on VOC2007 test set.}
	\label{table:Net_grid_search}
	\vspace{-1mm}
\end{table}  

\begin{figure*}[t]
	\begin{center}		
	\includegraphics[width=\linewidth]{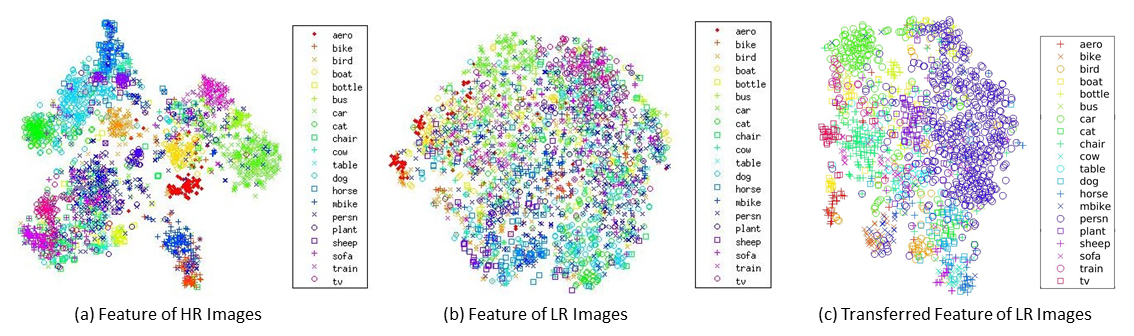}
	\end{center}
	\vspace{-8pt}
	\caption{The tSNE of features on VOC2007 test set. (a) Feature (2048-D) of High Resolution (HR) images, (b) feature (2048-D) of Low Resolution (LR) images, (c) transferred feature (100-D) of LR image.}
	\label{fig:VOC07_test_HR_LR_Transferred_tSNE}
\end{figure*}

\subsection{Feature Transfer Network}\label{sec:feature_feature_network}
The feature transfer network is shallow, with two fully connected layers. Let $N_1$ and $N_2$ denote the neurons of the first and second fully connected layers, respectively. We conduct grid search to find the optimal combination for $N_1$ and $N_2$, as shown in Table~\ref{table:Net_grid_search}. The number $N_2$ is determined by the number of clusters $k$ for the pseudo labels in k-means.

As we can see, when the neurons of $N_2$ is fixed, the mAP increases as the neurons of $N_1$ increases. This is because the capacity of the two-layers feature transfer network increases as the neurons increases in $N_1$. However, given a fixed number of neurons of $N_1$, the value of mAP would increase first, and then decrease when the value of neurons in $N_2$ is larger enough, maybe $200$ is a threshold value in our two-layer network as shown in the table. We observe that the hyperparameters with $N_2=100$ and $N_1=4096$ for the neurons give us the best performance. We use the same values in our experiment. 

\subsection{Low Resolution Image Classification}
We evaluate the performance of image classification in the context of binary classification task on the VOC2007 test set using SVM~\cite{chang2011libsvm} classifier in matlab. We have compared our algorithm with two baselines: Baseline-HR and Baseline-LR, discussed below. Baseline-HR is to use the extracted high resolution features (2048-D) of VOC2007 trainval set to train the SVM and report the classification performance on VOC2007 test set. It is similar for Baseline-LR, but with the extracted low resolution features (2048-D). Our method transfers the low resolution feature from 2048-D to 100-D. Therefore, we train the SVM using the 100-D features for each class. We show the comparison in Table~\ref{table:VOC07_Bbox_baseline}.


The Baseline-HR is the upper bound of our method, and Baseline-LR is the lower bound. As we can see from the Table~\ref{table:VOC07_Bbox_baseline}, the proposed unsupervised deep feature transfer is able to boost the low resolution image classification by about $2\%$. Except for the classes of ``bottle'' and ``sheep'', our method outperforms the Baseline-LR. As shown in Fig.~\ref{fig:VOC07_test_HR_LR_Transferred_tSNE}, we find the transferred low resolution features are separated much better than the extracted low resolution features. Those indicate that the proposed unsupervised deep feature transfer algorithm does help transfer more discriminative representations from high resolution features. Therefore, it boost on low resolution images classification task. The feature transfer network could also be embedded into the state-of-the-art deep neural networks as an plug-in module to enhance the learned features. 

\section{Conclusion}
In this paper, we propose an unsupervised deep feature transfer algorithm for low resolution image classification. The proposed two-layer feature transfer network is able to boost the classification by 2\% on mAP. It can be embedded into the state-of-the-art deep neural networks as a plug-in feature enhancement module. While our current experiments focus on generic classification, we expect our feature enhancement module to be very useful in detection, retrieval, and category discovery settings as well in the future.

\section*{Acknowledgment}
Dr. Zhang was supported by MERL. Mr. Wu and Prof. Wang were supported in part by NSF NRI and USDA NIFA under the award no. 2019-67021-28996 and KU General Research Fund (GRF).
{\small
\bibliographystyle{ieee}
\bibliography{small_object_iccv19_RLQ}
}

\end{document}